\documentclass{ws-procs9x6}

\begin{document}

\title{Deep Learning For Smile Recognition}

\author{Patrick O. Glauner}

\address{Interdisciplinary Centre for Security, Reliability and Trust, University of Luxembourg\\ 
2721 Luxembourg, Luxembourg \\
Email: patrick.glauner@uni.lu \\
snt.uni.lu \\
}

\begin{abstract}
Inspired by recent successes of deep learning in computer vision, we propose a novel application of deep convolutional neural networks to facial expression recognition, in particular smile recognition.
A smile recognition test accuracy of 99.45\% is achieved for the Denver Intensity of Spontaneous Facial Action (DISFA) database, significantly outperforming existing approaches based on hand-crafted features with accuracies ranging from 65.55\% to 79.67\%. The novelty of this approach includes a comprehensive model selection of the architecture parameters, allowing to find an appropriate architecture for each expression such as smile. This is feasible because all experiments were run on a Tesla K40c GPU, allowing a speedup of factor 10 over traditional computations on a CPU.
\end{abstract}


\bodymatter

\section{Introduction}
Neural networks are celebrating a comeback under the term "deep learning" for the last ten years by training many hidden layers allowing to self-learn complex feature hierarchies. This makes them of particular interest for computer vision, in which feature description is a long-standing issue.
Many advances have been reported in this period, including new training methods and a paradigm shift of training from CPUs to GPUs. As a result, those advances allow to train more reliable models much faster. This has for example resulted in breakthroughs\cite{hinton_speech} in signal processing. Nonetheless, deep neural networks are not a magic bullet and successful training is still heavily based on experimentation.

The Facial Action Coding System (FACS) \cite{FACS} is a system to taxonomize any facial expression of a human being by their appearance on the face. Action units describe muscles or muscle groups in the face, are set or unset and the activation may be on different intensity levels.
State-of-the art approaches in this field mostly rely on hand-crafted features leaving a lot of potential for higher accuracies. In contrast to other fields such as face or gesture recognition, only very few works on  deep learning applied to facial \textit{expression} recognition have been reported so far\cite{gudi} in which the architecture parameters are fixed. We are not aware of publications in which the architecture of a deep neural network for facial expression recognition is subject to extensive model selection. This allows to learn appropriate architectures per action unit.

\section{Deep neural networks}
Training neural networks is difficult, as their cost functions have many local minima. The more hidden layers, the more difficult the training of a neural network. Hence, training tends to converge to a local minimum, resulting in poor generalization of the network.
In order to overcome these issues, a variety of new concepts have been proposed in the literature, of which only a few can be named in this chapter. Unsupervised pre-training methods, such as autoencoders \cite{ng_tutorial} allow to initialize the weights well in order for backpropagation to quickly optimize them. The Rectified Linear Unit (ReLU) \cite{rectified} and dropout \cite{dropout_simple} are new regularization methods.
The new training methods and other new concepts can also lead to significant improvements of shallow neural networks with just a few hidden layers.
Convolutional neural networks (CNNs) were initially proposed by LeCun \cite{MNIST_LeNet} for the recognition of hand-written digits. A CNN consists of two layers: a convolutional layer, followed by a subsampling layer.
Inspired by biological processes and exploiting the fact that nearby pixels are strongly correlated, CNNs are relatively insensitive to small translations or rotations of the image input. 

Training deep neural networks is slow due to the number of parameters in the model. As the training can be described in a vectorized form, it is possible to massively parallelize it. Modern GPUs have thousands of cores and are therefore an ideal candidate for the execution of the training of neural networks. Significant speedups of factor 10 or higher\cite{tesla} have been reported.
A difficulty is to write GPU code. In the last few years, more abstract libraries have been released.

\section{DISFA database}
\label{chapter:stateoftheart}
\begin{figure}[h!]
    \centering
    \hfill
    \includegraphics[width=0.2\textwidth]{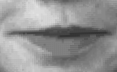}
    \hfill
    \includegraphics[width=0.2\textwidth]{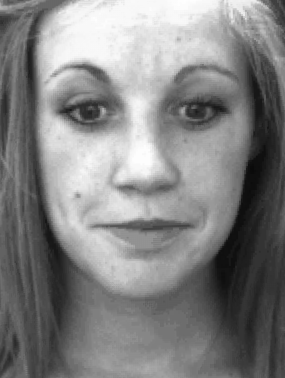}
    \hfill~
    \caption{Different input parts: a) mouth, b) face \cite{DISFA}. (Not at actual input size/proportions.)}
    \label{fig:mouth_face}
\end{figure}

The Denver Intensity of Spontaneous Facial Action (DISFA) \cite{DISFA} database consists of 27 videos of 4844 frames each, with 130,788 images in total. Action unit annotations are on different levels of intensity, which are ignored in the following experiments and action units are either set or unset. DISFA was selected from a wider range of databases popular in the field of facial expression recognition because of the high number of smiles, i.e. action unit 12. In detail, 30,792 have this action unit set, 82,176 images have some action unit(s) set and 48,612 images have no action unit(s) set at all. Fig.~\ref{fig:mouth_face} contains a sample image of DISFA.

In the original paper on DISFA \cite{DISFA} multi-class SVMs were trained for the different levels 0-5 of
action unit intensity. Test accuracies for the individual levels and for the binary action unit recognition
problem are reported for three different hand-crafted feature description techniques. In those three cases,
accuracies of 65.55\%, 72.94\% and 79.67\% for smile recognition are reported.

\section{Smile recognition}
\label{chapter:smile}
In the following experiments, an aligned version of DISFA is used. In this aligned version, the faces have been cropped and annotated with facial landmark points. Facial landmark points allow to compute a bounding box to fit the mouth in all images. In the experiments, two inputs are used: the mouth and face, downscaled to $85\times 69$ and $128\times 104$ pixels, respectively. Both inputs are used to assess if the mouth alone is as expressive as or even more expressive than the entire face for smile recognition.

\subsection{Model}
The architecture of the network is as follows:
The input images are fed into a convolution comprising a convolutional and a subsampling layer. That convolution may be followed by more convolutions to become gradually more invariant to distortions in the input. In the second stage, a regular neural network follows the convolutions in order to discriminate the features learned by the convolutions. The output layer consists of two units for smile or no smile.
The novelty of this approach is that the exact number of convolutions, number of hidden layers and size of hidden layers are not fixed but subject to extensive model selection in Sec.~\ref{chapter:param}.

\subsection{Experiment setting}
\label{chapter:setting}
Due to training time constraints, some parameters have been fixed to reasonable and empirical values, such as the size of convolutions ($5\times 5$ pixels, 32 feature maps) and the size of subsamplings ($2\times 2$ pixels using max pooling). All layers use ReLU units, except of softmax being used in the output layer. The learning rate is fixed to $\alpha = 0.01$ and not subject to model selection as it would significantly prolong the model selection. The same considerations apply to the momentum, which is fixed to $\mu = 0.9$.

The entire database has been randomly split into a 60\%/20\%/20\% training/validation/test ratio. Training neural networks comes with uncertainties, mostly due to the random initialization of the weights, but also due to that random split of the data. Evaluations have shown that for 10 similar experiments carried out, the standard deviation of the test accuracy is 0.041725\%. Because of this low standard deviation, performing each experiment exactly once has only a very low bias and is therefore relatively safe to do for reasons of faster training time.
Throughout the experiments, the classification rate is used as the accuracy measure. 

The model is implemented using Lasagne \cite{lasagne} and the generated CUDA code is executed on a Tesla K40c \cite{tesla} as training on a GPU allows to perform a comprehensive model selection in a feasible amount of time. Stochastic gradient descent with a batch size of 500 is used.

\subsection{Parameter optimization}
\label{chapter:param}
Table~\ref{table:cv_values} contains the four parameters to be optimized: the number of convolutions, the number of hidden layers, the number of units per hidden layer and the dropout factor. Each parameter was optimized independently due to training time constraints. This may not lead to an optimal model, but has proven to work empirically well. Each model was trained for 50 epochs in the model selection.

\begin{table}[h!]
\tbl{Parameters and possible values used in model selection.}
{\begin{tabular}{@{}ccc@{}}\toprule
Parameter & Values & Default value \\
\colrule
\#Convolutions & 1, 2, 3 & 1 \\
\#Hidden layers& 1, 2, 3 & 1 \\
\#Units / hidden layer & 100, 200, 300, 400 & 100 \\
Dropout & 0, 0.1, 0.5, 0.7 & 0.5 \\\botrule
\end{tabular}}
\label{table:cv_values}
\end{table}

For both inputs, Table~\ref{table:cv_both_final_models_full} contains the final models selected. For the mouth input, there is a preference to more convolutions and more hidden layers. This is the case because slight translations or rotations in the mouth input have stronger consequences on the classification result. In the entire face, that sort of distortions may be less of a problem because other parts of the face such as the cheeks contribute to smile recognition, too.

\begin{table}[h!]
\tbl{Selected parameter values for mouth and face input.}
{\begin{tabular}{@{}ccccc@{}}\toprule
Input & \#Convs & \#Hidden layers & \#Units / hidden layer & Dropout \\
\colrule
Mouth & 2 & 2 & 400 & 0.1 \\
Face & 1 & 1 & 400 & 0 \\\botrule
\end{tabular}}
\label{table:cv_both_final_models_full}
\end{table}

\subsection{Results and discussion}
Both final models were trained for 1000 epochs. The test accuracies of both models started to converge after about 300 epochs. For the mouth and face inputs, the best accuracies were achieved after 700 and 1000 epochs with 99.45\% and 99.34\%, respectively. Both models significantly outperform the state-of-the-art SVM baselines reported in Sec.~\ref{chapter:stateoftheart} ranging from 65.55\% to 79.67\%.
Overall, there is no strong preference for either the mouth or face input. Further experiments with a reduced dataset containing only 70\% of the images that have no action unit(s) set at all support this hypothesis. Concretely, the test accuracies for the mouth and face input reduced to 99.24\% and 99.26\%, respectively. Thus, the difference between the two models has been further reduced and this time giving a very low preference for the face input. Nonetheless, this difference is not representative as it is within the experiment error standard deviation reported in Sec.~\ref{chapter:setting}.

Training time per epoch are 82 seconds and 41 seconds for the mouth and face input models, respectively. Experiments have shown that the training time mostly depends on the number of convolutions. Using the Tesla K40c GPU has allowed to speed up the training time by factor ten over the use of a CPU to execute the CPU code generated by the library. This clearly demonstrates the importance of training on a GPU to do a comprehensive model selection in a feasible amount of time.

\section{Conclusions and future work}
Deep learning is an umbrella term for training neural networks with potentially many hidden layers using new training methods allowing to learn complex feature hierarchies from data. Applied to action unit recognition and smile recognition in particular, a deep convolutional neural network model with an overall accuracy of 99.45\% significantly outperforms existing approaches.
The underlying extensive model selection allows to find for each action unit an appropriate architecture in order to maximize test accuracies.
In the future, we will extend the model to images from multiple databases and to make predictions in image sequences. 

\bibliographystyle{ws-procs9x6}

\end{document}